\documentclass[10pt,twocolumn,letterpaper]{article}

\usepackage[cvprfinal]{cvpr}      % To produce the REVIEW version
\usepackage{graphicx}
\usepackage{amsmath}
\usepackage{amssymb}
\usepackage{booktabs}
\usepackage{caption}
\usepackage{float}
\usepackage{color}
\usepackage[dvipsnames]{xcolor}
\usepackage[dvipsnames]{xcolor}

\definecolor{cvprblue}{rgb}{0.21,0.49,0.74}
\usepackage[pagebackref,breaklinks,colorlinks,citecolor=cvprblue]{hyperref}

\def\paperID{940} % *** Enter the Paper ID here
\def\confName{CVPR}
\def\confYear{2024}

\title{FlowZero: Zero-Shot Text-to-Video Synthesis with \\ LLM-Driven Dynamic Scene Syntax}

\author{Yu Lu$^{1}$\quad Linchao Zhu$^{2}$\quad Hehe Fan$^{2}$\quad Yi Yang$^{2}$\\
$^{1}$ReLER Lab, University of Technology Sydney\\
$^{2}$CCAI, Zhejiang University\\
{\tt\small aniki.yulu@gmail.com}\\
}
\begin{document}
\twocolumn[{%
\maketitle

\vspace{-1cm}

\begin{figure}[H]
\hsize=\textwidth % cvpr 
\centering
\includegraphics[width=17cm]{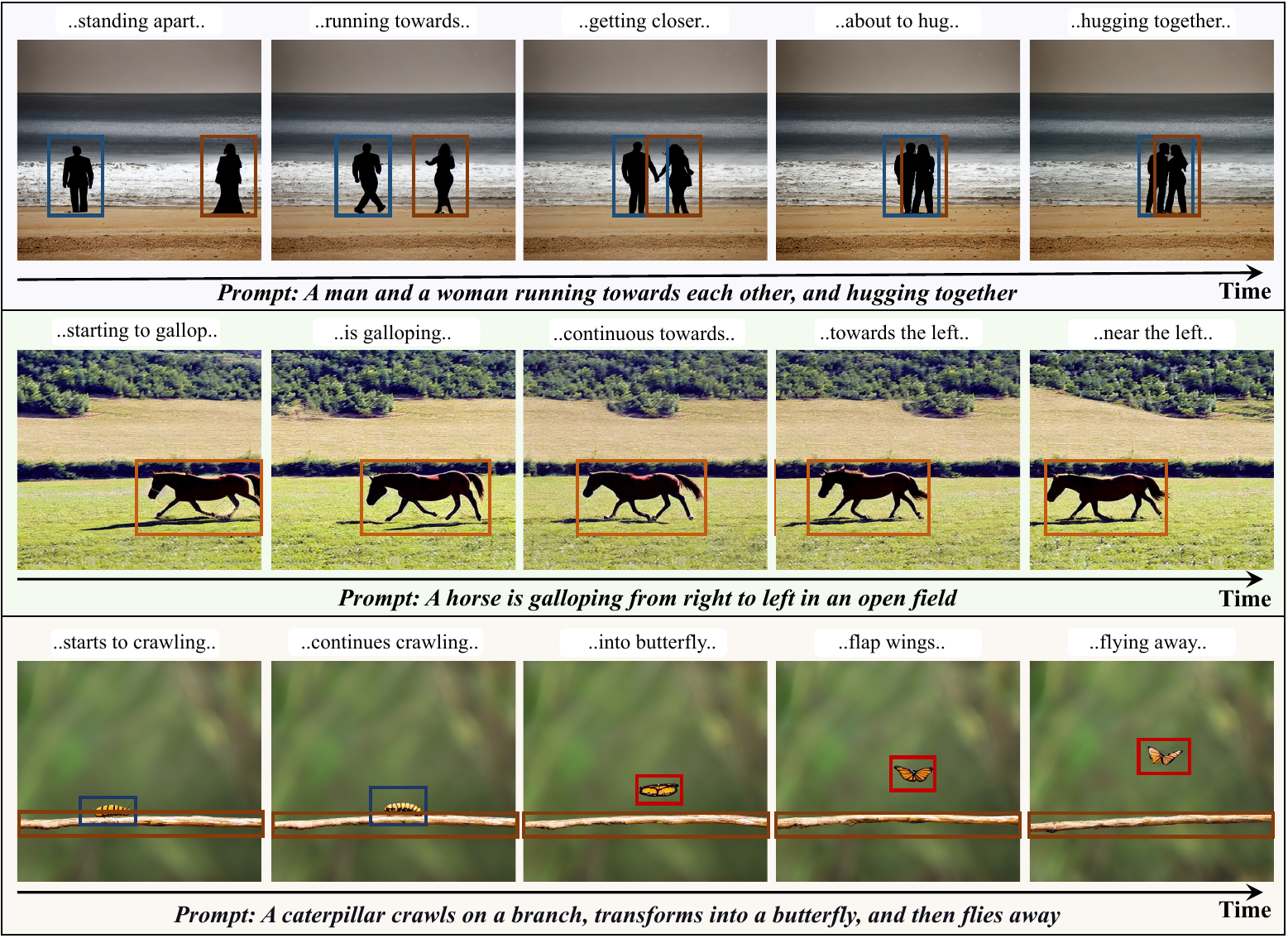}
\caption{\textbf{Zero-shot text-to-video generation.} We present a new framework for text-to-video generation with exceptional temporal coherence, featuring \textbf{realistic object movements, transformations, and background motion} within the generated videos.}
\end{figure}
\label{fig:teaser}
}]

\begin{abstract}

Text-to-video~(T2V) generation is a rapidly growing research area that aims to translate the scenes, objects, and actions within complex video text into a sequence of coherent visual frames.
We present FlowZero, a novel framework that combines Large Language Models (LLMs) with image diffusion models to generate temporally-coherent videos.
FlowZero uses LLMs to understand complex spatio-temporal dynamics from text, where LLMs can generate a comprehensive dynamic scene syntax~(DSS) containing scene descriptions, object layouts, and background motion patterns.
These elements in DSS are then used to guide the image diffusion model for video generation with smooth object motions and frame-to-frame coherence.
Moreover, FlowZero incorporates an iterative self-refinement process, enhancing the alignment between the spatio-temporal layouts and the textual prompts for the videos. 
To enhance global coherence, we propose enriching the initial noise of each frame with motion dynamics to control the background movement and camera motion adaptively.
By using spatio-temporal syntaxes to guide the diffusion process, FlowZero achieves improvement in zero-shot video synthesis, generating coherent videos with vivid motion.
Project page: \url{https://flowzero-video.github.io/}
\end{abstract}    
\section{Introduction}
\label{sec:intro}

In the field of AI-generated content, there has been growing interest in expanding the generative capabilities of pre-trained text-to-image (T2I) models to text-to-video (T2V) generation~\cite{animatediff,cogvideo,direct2v,free-bloom,ImagenVideo,MagicVideo,t2vz,videofusion,makeavideo}.
Recent studies have introduced zero-shot T2V~\cite{t2vz, free-bloom, direct2v}, which aims to adapt image diffusion models for video generation without additional training. These methods utilize the ability of image diffusion models, originally trained on static images, to generate frame sequences from video text prompts. 
However, generating coherent dynamic visual scenes in videos remains challenging due to the succinct and abstract nature of video text prompts.

Meanwhile, Large Language Models (LLMs) demonstrated their capability to generate layouts to control visual modules, especially image generation models~\cite{controlgpt,layoutgpt,lmd}.
These capabilities indicate a potential for LLMs to understand complex video prompts and generate fine-grained spatio-temporal layouts to guide video synthesis.
However, generating spatio-temporal layouts for videos is more intricate, necessitating the LLMs to comprehend and illustrate how objects move and transform over time.

Furthermore, recent research~\cite{free-bloom,direct2v} in zero-shot T2V proposes utilizing LLMs to break down video text into frame-level descriptions.
These descriptions are crafted to represent each moment or event within the video, guiding image diffusion models to generate semantic-coherent videos.
However, these frame-level descriptions only capture the basic temporal semantics of video prompts, lacking detailed spatio-temporal information necessary for ensuring smooth object motion and consistent frame-to-frame coherence in videos.
Additionally, representing global background movement to depict camera motion is crucial for immersive video generation~\cite{camera1,camera2}, which further complicates video generation.

In this paper, we introduce \textbf{\textit{FlowZero}}, a novel framework that integrates LLMs with image diffusion models to generate temporally-coherent videos from text prompts. 
FlowZero utilizes LLMs for comprehensive analysis and translating the video text prompt into a proposed structured Dynamic Scene Syntax~(DSS). Unlike previous methods that only provide basic semantic descriptions, the DSS contains scene descriptions, layouts for foreground objects, and background motion patterns.
Foreground layouts contain a series of bounding boxes that define each frame's spatial arrangement and track changes in the positions and sizes of objects. This ensures that the coherent object motion and transformation align with the textual prompt.
Additionally, FlowZero incorporates an iterative self-refinement process. This process effectively enhances the alignment between the generated layouts and the textual descriptions, specifically addressing inaccuracies such as spatial and temporal errors. In the self-refinement process, the generated layouts are iteratively compared and adjusted against the text through a feedback loop, ensuring a high fidelity and coherence of the spatio-temporal layouts. 

FlowZero prompts LLMs to predict background motion patterns to enhance temporal coherence and consistency, which can be used to control global scenes and camera motion in video frames.
For instance, consider a text that describes a horse running from right to left, as shown in the middle example of Figure~\textcolor{red}{1}. The LLMs predict a corresponding camera motion, making the background move from left to right, enhancing the video's immersiveness~\cite{camera1,camera2}.
The background motion pattern includes specific directions and speeds.
We introduce a motion-guided noise shifting (MNS) technique, shifting the initial noise of each frame according to the predicted background motion direction and speed, leading to smoother video synthesis.

FlowZero achieves a significant advancement in zero-shot text-to-video synthesis, utilizing the spatio-temporal planning ability of LLMs to generate detailed frame-by-frame syntax to enhance text-to-video generation. The fusion of these technologies within the FlowZero framework enables the generation of temporally-coherent, visually appealing videos directly from textual prompts.

Our contributions are summarised as follows:

\begin{itemize}
    \item We introduce FlowZero, which uses LLMs to convert text into Dynamic Scene Syntax, leading to accurate frame-by-frame video instructions. The framework's iterative self-refinement process ensures better alignment of spatio-temporal layouts with text prompts, enhancing video synthesis coherence and fidelity.

    \item The framework improves the global coherence of videos with adaptively controlled background motion through motion-guided noise shifting, increasing the realism of scene and camera motion.

\item Through extensive experiments and evaluations, we demonstrate FlowZero's capability to generate temporally-coherent videos that accurately depict complex motions and transformations as described in textual prompts.
\end{itemize}

\begin{figure*}[h]
\centering
   \setlength{\abovecaptionskip}{0.5cm}
   \includegraphics[scale = 0.52]{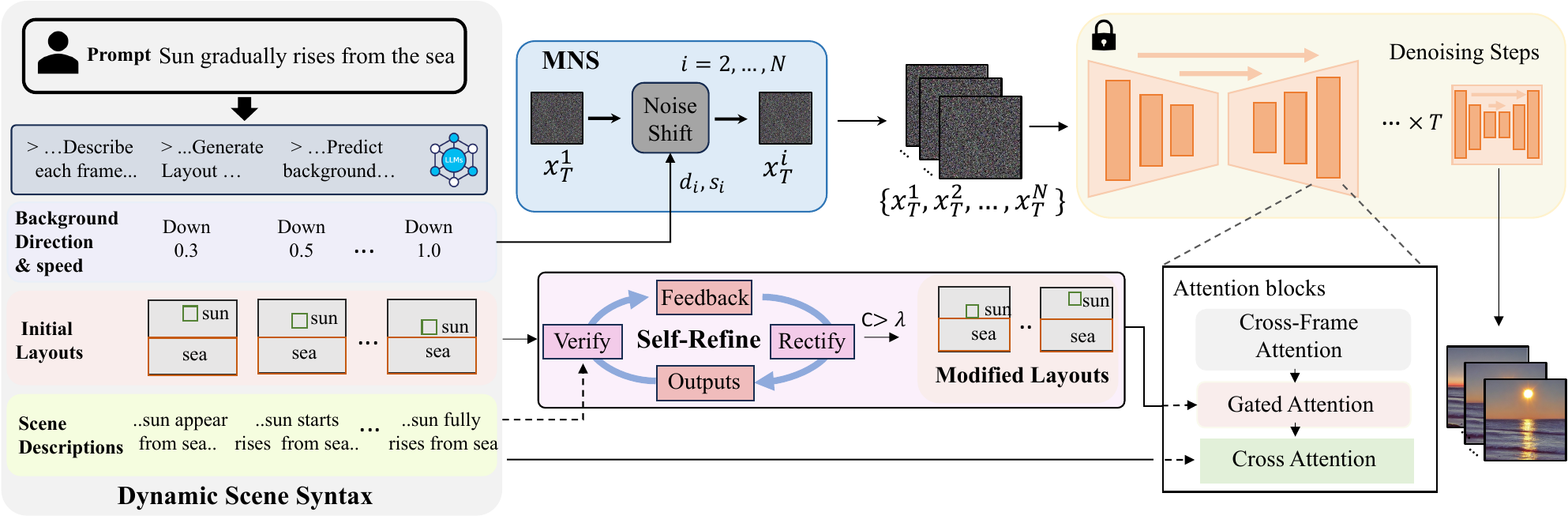}
     \caption{
     \textbf{Overview of FlowZero.} Starting from a video prompt, we first instruct the LLMs~(\ie, GPT4) to generate serial frame-by-frame syntax, including scene descriptions, foreground layouts, and background motion patterns. We employ an iterative self-refinement process to improve the generated spatio-temporal layouts. This process includes implementing a feedback loop where the LLM autonomously verifies and rectifies the spatial and temporal errors of the initial layouts. The loop continues until the confidence score 
    $C$ for the modified layouts exceeds a predefined threshold $\lambda$.
    Next, we perform motion-guided noise shifting ~(MNS) to obtain the initial noise for each frame $i$ by shifting the first noise with predicted background motion direcction~$d_{i}$ and speed~$s_{i}$. 
     Then, a U-Net with cross-attention, gated attention, and cross-frame attention is used to obtain $N$ coherent video frames.
     }
\vspace{-0.1cm}
   \label{fig:pipeline}
\vspace{-0.2cm}
\end{figure*}
\section{Related Work}
\label{sec:formatting}

\paragraph{Text-to-Video Generation.}
Text-to-video (T2V) generation has evolved from initial variational autoencoders~\cite{early1,early2} and GANs~\cite{gan} to advanced diffusion-based techniques~\cite{cogvideo,animatediff,lvdm,MagicVideo,makeavideo,free-bloom,direct2v,t2vz}, signifying a major advancement in synthesis methods. Although video diffusion models create high-quality visuals, training T2V models is often computationally expensive. This has led to the exploration of alternative approaches that balance efficiency and quality.
Recent advancements~\cite{t2vz,free-bloom,direct2v} have explored leveraging image diffusion models pre-trained on static images~\cite{lmd} to sidestep the demanding training process for T2V. For example, Text-to-Video Zero~\cite{t2vz} uses linear transformations and attention mechanisms to maintain video coherence. Free-Bloom~\cite{free-bloom} and DirecT2V~\cite{direct2v}, use Large Language Models~(LLMs) to guide image diffusion models with descriptive scene prompts for sequential frames.
However, these approaches struggle to capture the intricate object dynamics and background motion of videos, often leading to less expressive and coherent video generation. 
In contrast to previous methods that only provide semantic descriptions for each frame, FlowZero utilizes LLMs to reason a more comprehensive Dynamic Scene Syntax, delivering detailed, frame-by-frame guidance to enhance the temporal coherence and realism of T2V outputs.

\paragraph{Visual Planning with Large Language Models.}
Advancements in text-to-image synthesis show that using an intermediate representation, such as a layout or segmentation map, greatly improves the alignment between generated images and their text descriptions~\cite{gligen,controlnet}.
Various methodologies~\cite{layoutgpt,controlgpt,lmd,visorgpt} harness the vast world knowledge embedded in LLMs to craft spatial layouts to guide the image generation process. This has resulted in the creation of images with a reasonable spatial arrangement that closely matches the given textual prompts. For instance, LMD~\cite{lmd} introduces a novel, training-free approach, guiding a diffusion model with a unique controller to generate images based on layouts from LLMs. Similarly, LayoutGPT~\cite{layoutgpt} employs a program-guided strategy, adapting LLMs to cater to layout-driven visual planning across diverse fields. 
Differing from these methods, FlowZero explores the spatio-temporal planning ability of LLMs for temporally-coherent video generation.
\section{Method}

As shown in Figure~\ref{fig:pipeline}, FlowZero initially leverages LLMs (e.g., GPT-4) to process a video prompt $\mathcal{T}$, generating frame-to-frame scene descriptions, foreground layouts, and background motion patterns. 
Subsequently, a self-refinement step corrects inconsistencies between the layouts and prompts, such as misaligned movement directions.
The frame synthesis begins with a $x_{T}^{1}$ noise sampled from a Gaussian distribution. Then, we perform a motion-guided noise shifting~(MNS) to shift the noise to obtain initial noises~$\{x_{T}^{1},x_{T}^{2},...,x_{T}^{N}\}$, encoding background motion direction and speed into each frame. A modified U-Net with various attention mechanisms is employed to synthesize video frames. Finally, through DDIM sampling and a decoder, the final $N$ video frames $\{\mathcal{D}(x_{0}^{i})_{i=1}^{N}\} \in \mathbb{R}^{N \times H \times W \times 3}$ are generated.

\subsection{Dynamic Scene Syntax Generation}

In this stage, we aim to use the LLMs, \ie, GPT-4~\cite{gpt4} to convert textual prompts into structured syntaxes for guiding the generation of temporally-coherent videos. These syntaxes include frame-by-frame descriptions, foreground object layouts, and background motion patterns.

\begin{itemize}
    \item \textbf{Scene Descriptions:} 
    Videos often depict a series of continuous events, such as the sunrise, beginning with the ``lighting in the edge" and gradually ``rising from the horizon". 
    We propose using LLMs to break down the video text prompt into detailed frame descriptions to depict these events.
    Given a video text prompt~$\mathcal{T}$, we instruct the LLMs to segment this prompt into detailed scene descriptions~$\{\mathcal{T}_{1}, \mathcal{T}_{2},...,\mathcal{T}_{N}\}$. These descriptions maintain consistent linguistic structures, ensuring that each prompt accurately conveys the visual content in a detailed manner.
    By providing a description for each frame, we can capture the temporal semantics of the video prompt.
            
    \item \textbf{Foreground Layout:}
    While scene descriptions provide semantic details for each frame, these high-level constraints are not sufficient to accurately depict specific object motion and transformations. To achieve coherent object motion, we prompt LLMs to generate a sequence of frame-specific layouts $\{{L_{1}, L_{2}, \ldots, L_{N}}\}$ that outline the spatial arrangement of foreground entities in each frame. These layouts are comprised of bounding boxes that define the position and size of the prompt-referenced objects, using the format: \( object: \{x_{1}, y_{1}, x_{2}, y_{2}\} \). Here, $object$ represents the category of the object along with any relevant attributes (for example, ``red car"), \( (x_{1}, y_{1}) \) and \( (x_{2}, y_{2}) \) denote the coordinates for the top-left and bottom-right vertices of the bounding box. These layouts provide more fine-grained conditions to ensure the foreground objects adhere to the visual and spatio-temporal cues the text provides.
                
    \item \textbf{Background Motion:} 
    Background motion plays a crucial role in enhancing the global coherence of videos, especially when dynamic foreground objects are involved. For example, in a video showing a horse running to the left, synchronizing the camera motion with the horse's direction can create a visually smooth effect, making the video more immersive and engaging~\cite{camera1, camera2}.
    To effectively simulate this, we first categorize potential background motion into eight moving directions: \{left, right, up, down, left\_up, left\_down, right\_up, right\_down\}, and include a ``random" option for non-directional movement. We also define a motion speed that ranges from 0 (no movement) to 1.0 (rapid movement).
    We use LLMs to determine the most appropriate background motion direction and speed for each frame. This helps us align it with the foreground movements as described in the scene. 
    By integrating background motions, we ensure global coherence and consistency in video sequences.
    \end{itemize}

Based on previous studies~\cite{layoutgpt, visorgpt, controlgpt,free-bloom,direct2v}, we instruct LLMs to generate these syntaxes through direct commands. For example, we use prompts like ``describe each frame" to create descriptions and ``generate layouts for each scene" to generate foreground layouts. We provide an example in context to enhance the stability and effectiveness of LLMs.

    \paragraph{Iterative Self-Refinement.}
    Due to the complex nature of reasoning in spatio-temporal dynamics, there may be discrepancies between the generated spatio-temporal layouts and the textual prompts. 
    As illustrated in Figure~\ref{fig:pipeline}, the sun initially moves downward over time, which contradicts the video prompt ``sun gradually rises".
    Previous research has shown that LLMs can verify and correct generated texts or codes~\cite{verify1,verify2,verify3,verify4}. Inspired by this, we propose an iterative self-refinement process to address potential misalignments between the initial spatio-temporal layouts \(\{L_{1}, L_{2}, \ldots, L_{N}\}\) and the textual prompts \(\{\mathcal{T}_{1}, \mathcal{T}_{2}, \ldots, \mathcal{T}_{N}\}\).
    The initial step of self-refinement involves prompting LLMs to verify spatial and temporal consistency between scene descriptions and layouts and provide detailed feedback. This feedback includes an analysis of problems, specific suggestions~(\eg, the sun should rise up instead of going down), and a confidence score $c$ from 1 to 5 to measure the alignment of layouts with descriptions.
    We found that providing numerically supported analysis and suggestions enhances the effectiveness of the self-refinement process. For example, examining and comparing particular coordinates of the bounding boxes can be particularly helpful. 
    We include in-context examples that clearly demonstrate the type of feedback most helpful for generating specific suggestions.
    Then, we prompt LLMs again to correct the layouts in the rectification step to improve spatial and temporal alignment with the textual prompts. This refinement process consists of multiple iterations, with the LLMs verifying and rectifying the layouts based on the previous iteration, leading to convergence toward an optimal layout representation. 
    The iterations continue until the confidence score $c$ is higher than a predefined alignment threshold $\lambda$. 

\subsection{Video Synthesis from Dynamic Scene Syntax}\label{sec:sythesis}

In this section, we seek to generate coherent video frames based on the generated DSS. 
As shown in the right part of Figure~\ref{fig:pipeline}, beginning with a noise~$x_{T}^{1}$ from standard Gaussian distribution, we first conduct Motion-guided noise shifting to obtain initial noises~$\{x_{T}^{1}, x_{T}^{2}, x_{T}^{3}, ..., x_{T}^{N}\}$ for each frame.
These noises are obtained by shifting noise~$x_{T}^{1}$ to match the background motion direction and speed predicted in the DSS generation stage. 
We will provide detailed information on Motion-guided noise shifting below.
We employ a modified U-Net with cross attention, gated attention, and cross-frame attention mechanisms~\cite{t2vz,direct2v,free-bloom}.
The cross-attention mechanism within the U-Net is designed to input scene descriptions, enabling the capture of diverse semantics for each frame. 
Simultaneously, the gated attention~\cite{gligen} inputs foreground layouts into the U-Net, managing the arrangement of objects across different frames.
We then convert the self-attention layer in the U-Net of the image diffusion model into cross-frame attention~\cite{free-bloom, direct2v, t2vz}, which performs attention between the query frame and previous frames.

\paragraph{Motion-guided Noise Shifting.}

Previous method~\cite{t2vz} performs a linear transformation on initial noises with fixed direction and speed to model global motion dynamics in video frames. 
In contrast, our approach allows LLMs to predict the background motion direction and speed adaptively for transforming noises, thereby significantly enhancing the global temporal coherence of videos.

Given the predicted background motion~$d$ and speed~$s$, a straightforward method is directly shifting the noise spatially for each frame. However, this often results in abrupt changes in low-level visual effects, such as color and lighting alterations in the video frames. To address this problem, we propose a technique to shift the phase of noises in the frequency domain~\cite{book1}. This method preserves the amplitude component to maintain low-level visual effects while modulating the phase component to simulate spatial noise shifting~\cite{phase1,phase2,phases3,phase4}, achieves smoother video frames.

Specifically, for each frame~$i$, we use the predicted background motion direction (\(d_{i}\)) and speed (\(s_{i}\)) to guide the spatial shift of noise. This is achieved by modulating the phase component of noise $x_{T}^{1}$ in the frequency domain, $T$ means the total diffusion step. The mathematical formulation is as follows:
\begin{equation}
x_T^{i} = \mathcal{F}^{-1}\left(\mathcal{F}(x^{1}_{T}) \cdot e^{-j \cdot 2\pi \cdot (i \cdot s_{i}) \cdot (d_{y}f_{y}+d_{x}f_{x})}\right)\nonumber,
\end{equation}
where \(\mathcal{F}\) and \(\mathcal{F}^{-1}\) denote the Discrete Fourier Transform (DFT) and its inverse, respectively. The frequencies in the y and x dimensions are represented by \(f_{y}\) and \(f_{x}\). The direction multipliers \(d_{y}\) and \(d_{x}\) are derived based on the motion direction~$d_{i}$. For instance, if the direction is ``left," \(d_{x}\) and \(d_{y}\) would be set to \{0,1\}. When the background motion direction remains the same across frames, the index will increase linearly, resulting in smooth motion effects.

In scenarios with non-directional movements ($d=random$), such as ``a goldfish swimming in a fish bowl," setting a static background for all frames can be unrealistic. Our method addresses this by adding random disturbances at the phase components of all frequencies, simulating natural scene variability, and enhancing video realism. 

We perform the noise-shifting technique in the frequency domain has several advantages: 

\begin{enumerate}
    \item Our method allows for easy modification of moving directions by adjusting the direction multipliers, offering greater flexibility than direct space shifting.
    \item Since our technique operates in the frequency domain, it is more efficient and computationally less intensive than spatial domain transformations, particularly for handling high-resolution and long videos.
    \item We can simulate realistic motionless scenes by adding random disturbances to the noises. 
\end{enumerate}

\section{Experiments}

\begin{figure*}[h]
\centering
   \setlength{\abovecaptionskip}{0.5cm}
   \includegraphics[scale = 0.34]{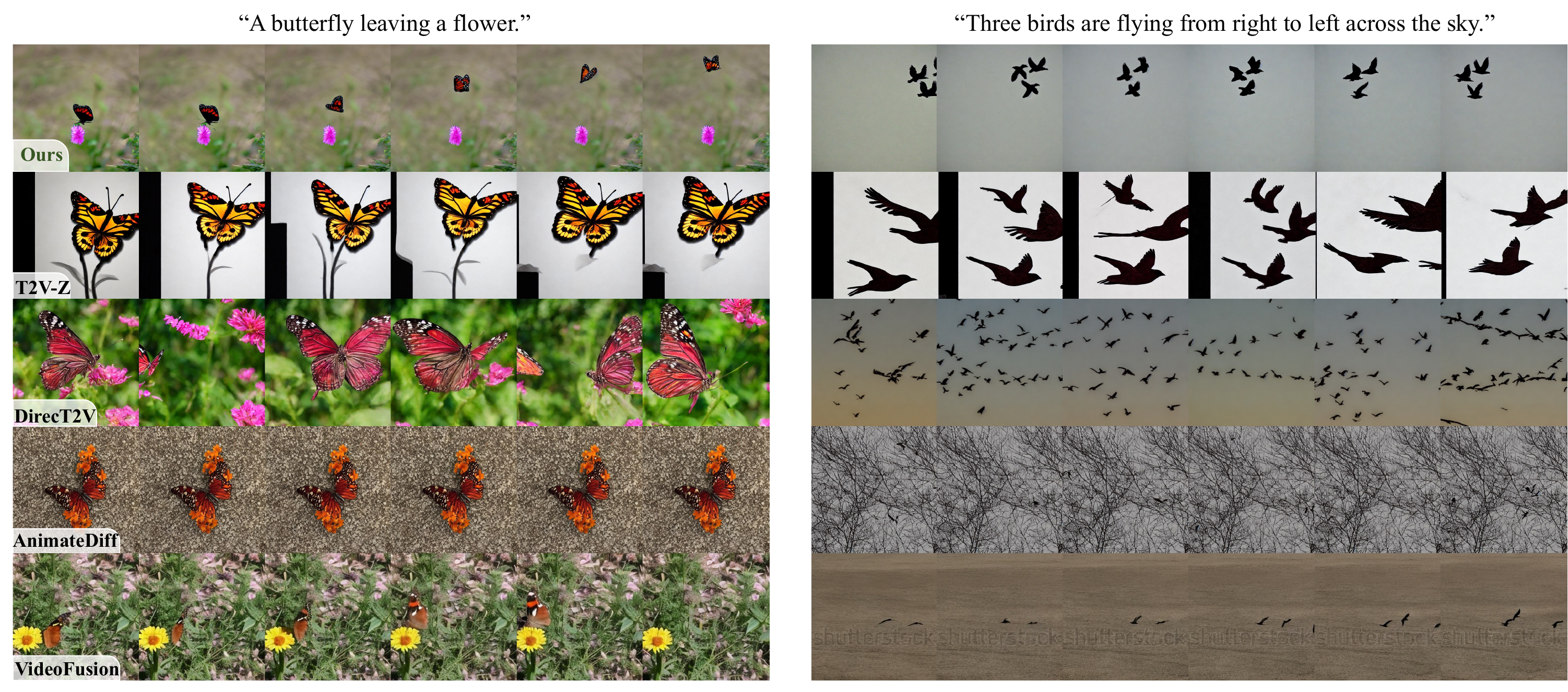}
\vspace{-0.2cm}
     \caption{\textbf{Qualitative comparison.} Our method can capture detailed object motion to generate temporally coherent frame sequences.
     }
\vspace{-0.1cm}
   \label{fig:compare1}
\end{figure*}

\subsection{Implementation Details}
We utilize the GLIGEN~\cite{gligen} as the base image diffusion model, which is pre-trained to generate images adhering to a layout.
We employ GPT-4~\cite{gpt4} to reason Dynamic Scene Syntax~(DSS). In our tests, we generate $N$ = 8 frames per video, each with a resolution of 512 × 512. 
However, our framework allows for generating any desired number of frames by instructing LLMs and increasing $N$. 
We set a threshold $\lambda$ of self-refinement as 3 and the maximum iteration as 5.
All experiments are conducted on a single NVIDIA V100 GPU.

\subsection{Comparisons with Baseline Methods}

\paragraph{Qualitative Comparison}

\begin{figure*}[h]
\centering
   \setlength{\abovecaptionskip}{0.5cm}
   \includegraphics[scale = 0.34]{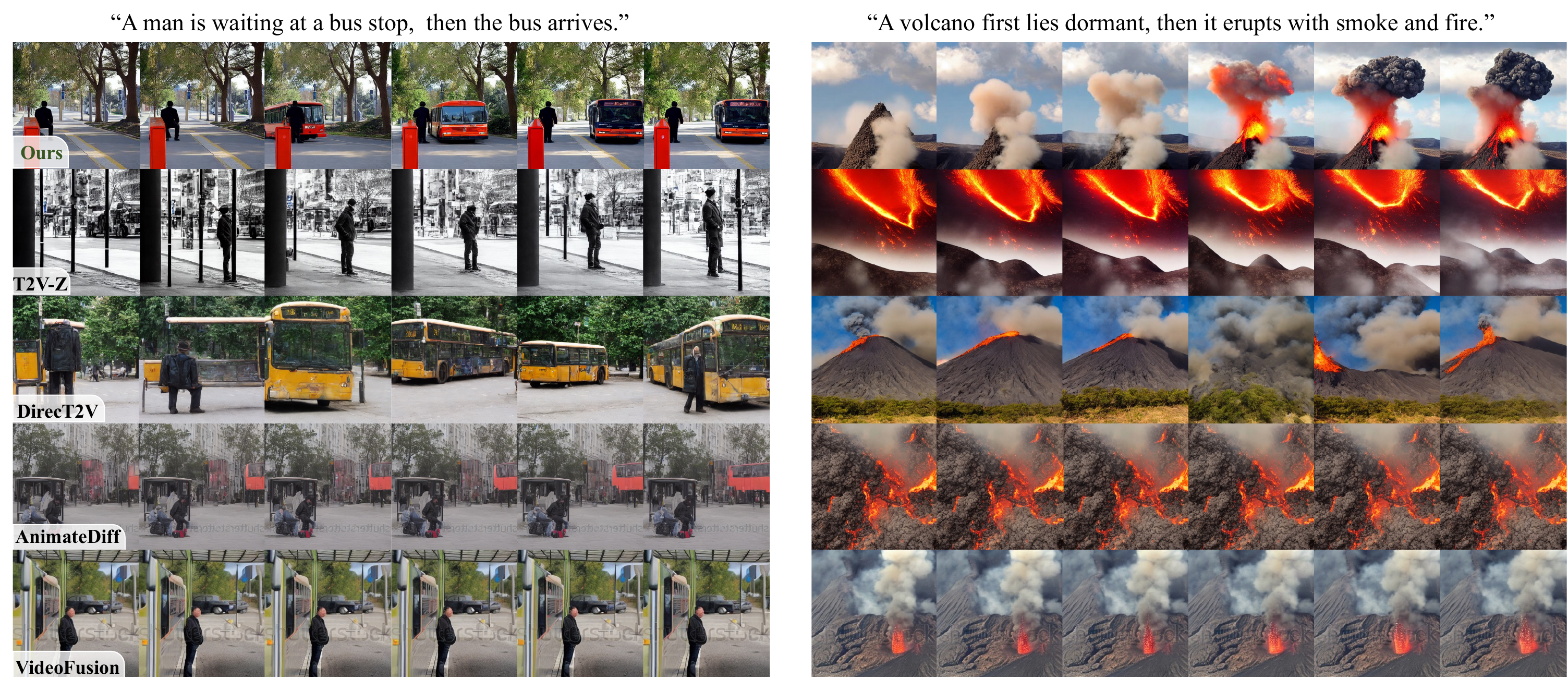}
\vspace{-0.2cm}
     \caption{\textbf{Qualitative comparison.} Our method can model intricate object transformations representing narrative structures in the video prompt.
     }
\vspace{-0.3cm}
   \label{fig:compare2}
\end{figure*}

\begin{table*}
    \centering
    \small
    \caption{\textbf{Quantitative Results.} We perform automatic metrics, \,  i.e., CLIP score and user study, to validate the effectiveness.}
    \begin{tabular}{lcccccc}
        \toprule
        & & Automatic Metric & \multicolumn{3}{c}{User Study} \\
         \cmidrule(lr){3-3} \cmidrule(lr){4-7}
        Method & Training-Free & CLIP Score$\uparrow$ & Semantic$\uparrow$ & Temporal$\uparrow$ & Quality$\uparrow$ & Rank$\downarrow$ \\
        \midrule
        AnimateDiff~\cite{animatediff} &  & 0.244  & 3.15 & 2.75 & 2.97 & 3.42 \\
        VideoFusion~\cite{videofusion} &  & 0.264 & 3.38 & 2.92 & 3.11 & 3.17 \\
        \midrule
        T2V-Z~\cite{t2vz} & \checkmark & 0.245  & 3.29 & 2.99 & 3.03 & 3.19 \\
        DirecT2V~\cite{direct2v} & \checkmark & 0.244  & 3.39 & 3.29 & 2.52 & 2.97 \\
        Ours & \checkmark & \textbf{0.267} &  \textbf{4.57} & \textbf{4.58} & \textbf{4.40} & \textbf{2.00} \\
        \bottomrule
    \end{tabular}
    \label{tab:compare}
\end{table*}

\begin{figure*}[h]
\centering
   \setlength{\abovecaptionskip}{0.5cm}
   \includegraphics[scale = 0.68]{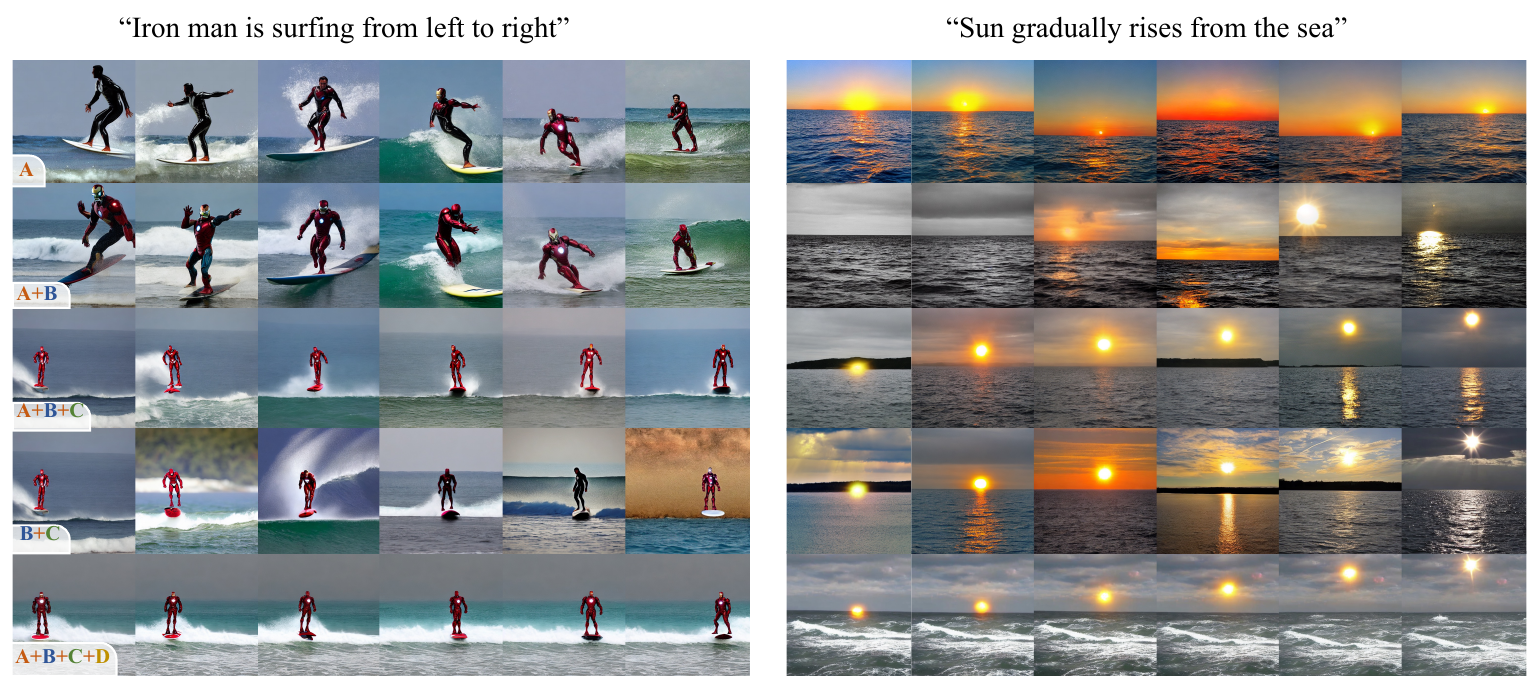}
     \caption{Ablation studies of the effectiveness of FlowZero. \textcolor{Bittersweet}{(\textbf{A}) cross-frame attention},  \textcolor{BlueViolet}{(\textbf{B}) scene descriptions},  \textcolor{OliveGreen}{(\textbf{C}) foreground layouts},  \textcolor{YellowOrange}{(\textbf{D}) motion-guided noise shifting}.
     }
\vspace{-0.2cm}
   \label{fig:ablation_blueprint}
\vspace{-0.3cm}
\end{figure*}

\begin{figure*}[h]
\centering
   \setlength{\abovecaptionskip}{0.5cm}
   \includegraphics[scale = 0.45]{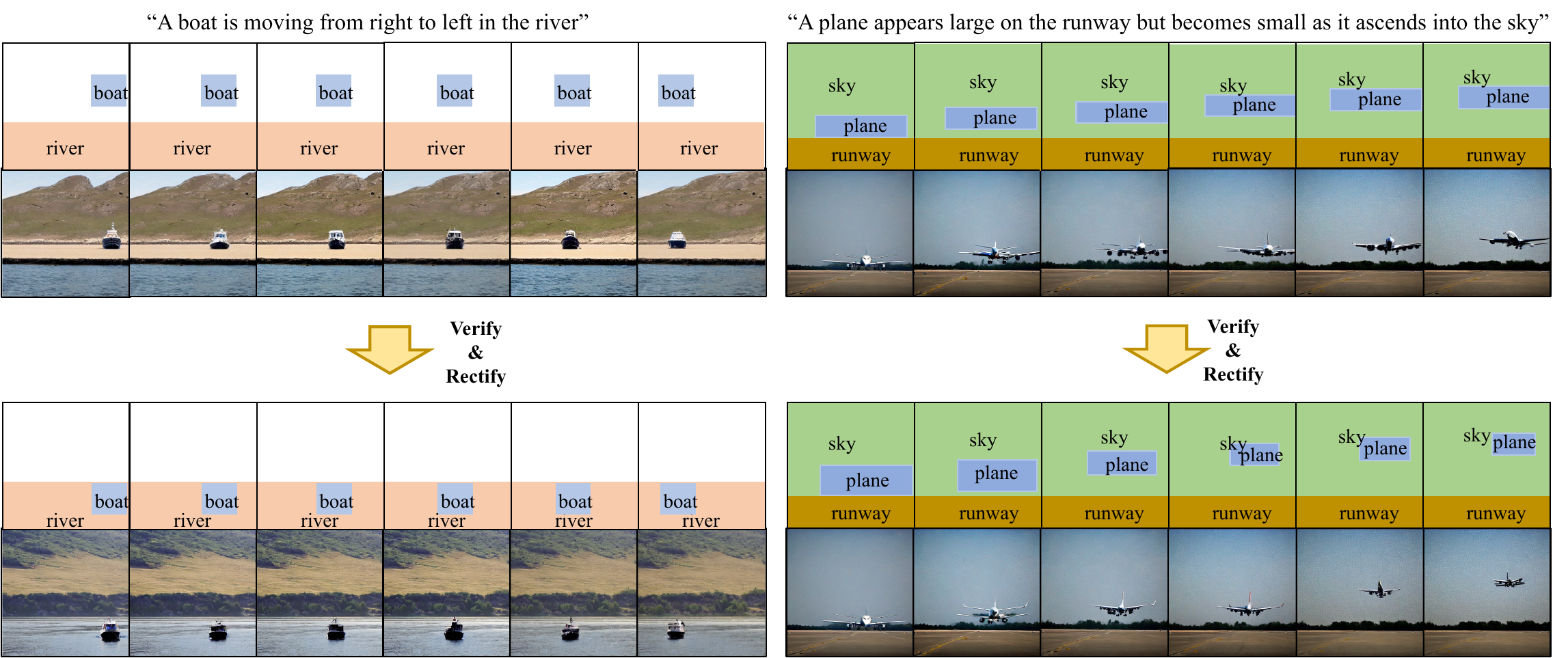}
     \caption{Analysis of the self-refinement process. The self-refinement mechanism verifies and rectifies spatial and temporal alignment between layouts and video prompts.
     }
   \label{fig:verify}
\vspace{-0.1cm}
\end{figure*}

In our qualitative comparative analysis, we compare videos generated using our FlowZero with several benchmark methods: zero-shot based methods, T2V-Z~\cite{t2vz}, DirecT2V~\cite{direct2v}, and training-based methods AnimateDiff~\cite{animatediff}, VideoFusion~\cite{videofusion}. We assess the performance in three scenarios: basic object motion rendering, multiple object motion depiction, and complex object transformations.

In our initial assessment, shown in Figure~\ref{fig:compare1}, we analyze videos generated from prompts featuring basic object motion. FlowZero effectively demonstrates the ability to depict smooth object motion, particularly showcasing a butterfly's departure from a flower.
However, other zero-shot techniques, such as T2V-Z and DirecT2V, only capture temporal semantics and struggle to model the coherent object motion. AnimateDiff and VideoFusion were trained on extensive video-text data~\cite{webvid2m} and exhibit temporal frame coherence. However, they fall short in rendering nuanced motion details, resulting in slightly stilted animations.
In scenarios involving multiple objects with designated movements, also presented in the right of Figure~\ref{fig:compare1}, FlowZero continues to excel, accurately animating specific objects and motion defined in the text prompts. Other methods struggle to precisely replicate the specified objects and movements, often resulting in a less accurate portrayal.
In Figure~\ref{fig:compare2}, we compare all methods of generating videos from prompts that describe complex object transformations. FlowZero distinguishes itself by vividly rendering transformations, such as a tranquil volcano erupting. Other methods do not effectively translate these temporal dynamics, leading to less coherent visual transformations.

By utilizing LLMs to plan the spatio-temporal syntax as guidance for the diffusion model, FlowZero surpasses other methods in text-prompted video generation. Its superior performance is particularly noticeable in the accurate motion of multiple objects and intricate object transformations.

\paragraph{Quantitative Comparison}

As shown in Table~\ref{tab:compare}, we first compare our methods with other four baseline methods, \ie, AnimateDiff~\cite{animatediff}, VideoFusion~\cite{videofusion}, T2V-Z~\cite{t2vz}, DirecT2V~\cite{direct2v} using CLIP score metrics~\cite{clip}. The CLIP metrics measure the semantic similarity between the text and video frames. Our method achieves the highest performance by prompting LLMs to deduce more semantics from both spatial and temporal dimensions.

 Due to the complexity of quantitatively evaluating videos with intricate temporal dynamics, we conducted a user study to validate the effectiveness of our method.
 We recruited 20 people from academia and industry to conduct this survey.
 We ask users to provide feedback on the semantic accuracy, temporal coherence, and video quality of the videos generated by five different methods. It is evident that users prefer our method and achieve better results, surpassing even training-based methods, \eg, AnimateDiff~\cite{animatediff} and VideoFusio~\cite{videofusion}.
Furthermore, our methods achieve significant improvement over other zero-shot methods, \eg, T2V-Zero~\cite{t2vz} and DirecT2V~\cite{direct2v} on temporal coherence, which validates the effectiveness of our approach.
% The feedback from users confirms the superiority of our approach in generating temporally coherent, faithful, and diverse videos.

\subsection{Ablation Study}

\paragraph{Effectiveness of FlowZero}

In Figure~\ref{fig:ablation_blueprint}, we conduct a comprehensive ablation study to validate the effectiveness of key components in FlowZero. This includes cross-frame attention, scene descriptions, foreground object layouts, and background motion for noise shifting. We begin with a baseline model that employs cross-frame attention to adapt U-Net, feeding it original video prompts alongside independent random noise for each frame. 
Row \#1 of Figure~\ref{fig:ablation_blueprint} demonstrates that the baseline generates videos with basic semantics like Ironman and surfing but fails to capture detailed object motion.
In row\#2, we replace the video prompt with generated scene descriptions for each frame, similar to previous methods Free-Bloom~\cite{free-bloom} and DirecT2V~\cite{direct2v}. However, we found that merely using temporal semantics resulted in a lack of coherent object motion, resulting in inconsistencies across frames.
Instead, in row\#3, by adding the layout to constrain the arrangement of foreground objects, we can clearly capture the coherent motion of the main object, such as ``from left to right" and ``rises". 
However, the video frames still display temporal inconsistency, such as in color, lighting, and global scene. 
In row\#4, we experimented with removing the cross-frame attention from U-Net, which means relying solely on a pure image diffusion model~\cite{gligen}. This modification resulted in a lack of inconsistencies in the representation of objects and backgrounds across frames, even though the layouts guide the object motion. 
Finally, by utilizing our motion-guided noise shifting technique in row\#5, we can smoothly control the motion direction and speed in the background, resulting in a coherent global scene.

\paragraph{Effectiveness of Self-Refinement Process}

We present two examples in Figure~\ref{fig:verify} to illustrate the effectiveness of our self-refinement process in correcting spatial and temporal errors of initial layouts.
In the first example, the video prompt describes ``a boat moving in the river." However, the initial generated layout incorrectly places the boat above the river. The boat is correctly positioned within the river through self-refinement, aligning with the prompt's spatial arrangement.
The second example describes a plane ascending into the sky. Initially, the size of the plane remains constant across all layouts over time, contradicting the expectation that it should appear smaller as it ascends. After refinement, the size of the plane decreases in later frames, accurately reflecting the prompt's temporal dynamics.

To quantitatively evaluate the effectiveness of the self-refinement process, we propose a benchmark comprising four spatio-temporal layout generation tasks. 
These tasks include multiple objects, object movements (left, right, up, down), size changes (big to small or small to big), and visibility variations (half or quarter visibility).
Each task includes 20 programmatically generated prompts, assessed using a rule-based metric. For instance, we calculate the change in object area across frames to evaluate size changes.
The results are displayed in Table~\ref{tab:self-refine}. 
We observe LLMs initially struggle to generate precise results that accurately reflect specific temporal changes, including object movement, size variation, and visibility. 
Moreover, through our self-refinement process, we noted a notable improvement in accuracy, particularly in tasks temporal visibility~(from 61\% to 78\%). The self-refinement mechanism consistently enhances spatial-temporal layout generation, effectively aligning the generated content with specific temporal requirements. 
These experiments confirm the effectiveness of our self-refinement process in improving the spatial-temporal coherence of the generated scenes.
\begin{table}
    \centering
    \small
    \caption{Quantitative analysis of the self-refinement process.}
    \begin{tabular}{lcccc}
        \toprule
        Method & Objects$\uparrow$ & Movement$\uparrow$ & Size$\uparrow$ & Visibility$\uparrow$ \\
        \midrule
        % LLaMA-2 & - &  - & - \\
        % GPT-3.5 & - &  - & - \\
        w/o self-refine & 90\% & 83\% & 80\% & 61\%\\
        w/ self-refine & 96\% &  93\% & 93\% & 78\% \\
        \bottomrule
    \end{tabular}
    \label{tab:self-refine}
\vspace{-0.1cm}
\end{table}

\section{Conclusion}

In this paper, we have investigated leveraging the spatial-temporal planning ability of Large Language Models to guide temporally-coherent text-to-video generation with image diffusion models. We prompt LLMs to generate comprehensive Dynamic Scene Syntax, including scene descriptions, layouts for foreground objects, and background motion patterns. The foreground layouts ensure coherent object motions and object transformations described in the prompt. Furthermore, the introduced iterative self-refinement can enhance the alignment between the generated spatio-temporal layouts and the textual descriptions, specifically addressing inaccuracies such as spatial and temporal errors. The background motion can be controlled by motion-guided noise shifting, leading to smoother video synthesis and a coherent global scene. We have performed extensive qualitative and quantitative experiments along with ablation studies to validate the effectiveness of our FlowZero framework. These experiments validate that FlowZero can generate temporally-coherent videos from complex video prompts.
{
    \small
    \bibliographystyle{ieeenat_fullname}
    \bibliography{main}

\begin{thebibliography}{33}
\providecommand{\natexlab}[1]{#1}
\providecommand{\url}[1]{\texttt{#1}}
\expandafter\ifx\csname urlstyle\endcsname\relax
  \providecommand{\doi}[1]{doi: #1}\else
  \providecommand{\doi}{doi: \begingroup \urlstyle{rm}\Url}\fi

\bibitem[Bain et~al.(2021)Bain, Nagrani, Varol, and Zisserman]{webvid2m}
Max Bain, Arsha Nagrani, G{\"{u}}l Varol, and Andrew Zisserman.
\newblock Frozen in time: {A} joint video and image encoder for end-to-end retrieval.
\newblock In \emph{2021 {IEEE/CVF} International Conference on Computer Vision, {ICCV} 2021, Montreal, QC, Canada, October 10-17, 2021}, pages 1708--1718. {IEEE}, 2021.

\bibitem[Dhuliawala et~al.(2023)Dhuliawala, Komeili, Xu, Raileanu, Li, Celikyilmaz, and Weston]{verify2}
Shehzaad Dhuliawala, Mojtaba Komeili, Jing Xu, Roberta Raileanu, Xian Li, Asli Celikyilmaz, and Jason Weston.
\newblock Chain-of-verification reduces hallucination in large language models.
\newblock \emph{arXiv preprint arXiv:2309.11495}, 2023.

\bibitem[Feng et~al.(2023)Feng, Zhu, Fu, Jampani, Akula, He, Basu, Wang, and Wang]{layoutgpt}
Weixi Feng, Wanrong Zhu, Tsu{-}Jui Fu, Varun Jampani, Arjun~R. Akula, Xuehai He, Sugato Basu, Xin~Eric Wang, and William~Yang Wang.
\newblock Layoutgpt: Compositional visual planning and generation with large language models.
\newblock \emph{CoRR}, abs/2305.15393, 2023.

\bibitem[Goodfellow et~al.(2014)Goodfellow, Pouget{-}Abadie, Mirza, Xu, Warde{-}Farley, Ozair, Courville, and Bengio]{gan}
Ian~J. Goodfellow, Jean Pouget{-}Abadie, Mehdi Mirza, Bing Xu, David Warde{-}Farley, Sherjil Ozair, Aaron~C. Courville, and Yoshua Bengio.
\newblock Generative adversarial networks.
\newblock \emph{CoRR}, abs/1406.2661, 2014.

\bibitem[Guo et~al.(2023)Guo, Yang, Rao, Wang, Qiao, Lin, and Dai]{animatediff}
Yuwei Guo, Ceyuan Yang, Anyi Rao, Yaohui Wang, Yu Qiao, Dahua Lin, and Bo Dai.
\newblock Animatediff: Animate your personalized text-to-image diffusion models without specific tuning.
\newblock \emph{CoRR}, abs/2307.04725, 2023.

\bibitem[Hansen and Hess(2007)]{phase1}
Bruce~C Hansen and Robert~F Hess.
\newblock Structural sparseness and spatial phase alignment in natural scenes.
\newblock \emph{JOSA A}, 24\penalty0 (7):\penalty0 1873--1885, 2007.

\bibitem[He et~al.(2022)He, Yang, Zhang, Shan, and Chen]{lvdm}
Yingqing He, Tianyu Yang, Yong Zhang, Ying Shan, and Qifeng Chen.
\newblock Latent video diffusion models for high-fidelity video generation with arbitrary lengths.
\newblock \emph{arXiv preprint arXiv:2211.13221}, 2022.

\bibitem[Heimann et~al.(2019)Heimann, Uithol, Calbi, Umilt{\`a}, Guerra, Fingerhut, and Gallese]{camera2}
Katrin Heimann, Sebo Uithol, Marta Calbi, Maria~Alessandra Umilt{\`a}, Michele Guerra, Joerg Fingerhut, and Vittorio Gallese.
\newblock Embodying the camera: An eeg study on the effect of camera movements on film spectators sensorimotor cortex activation.
\newblock \emph{PloS one}, 14\penalty0 (3):\penalty0 e0211026, 2019.

\bibitem[Ho et~al.(2022)Ho, Chan, Saharia, Whang, Gao, Gritsenko, Kingma, Poole, Norouzi, Fleet, and Salimans]{ImagenVideo}
Jonathan Ho, William Chan, Chitwan Saharia, Jay Whang, Ruiqi Gao, Alexey~A. Gritsenko, Diederik~P. Kingma, Ben Poole, Mohammad Norouzi, David~J. Fleet, and Tim Salimans.
\newblock Imagen video: High definition video generation with diffusion models.
\newblock \emph{CoRR}, abs/2210.02303, 2022.

\bibitem[Hong et~al.(2023{\natexlab{a}})Hong, Seo, Hong, Shin, and Kim]{direct2v}
Susung Hong, Junyoung Seo, Sunghwan Hong, Heeseong Shin, and Seungryong Kim.
\newblock Large language models are frame-level directors for zero-shot text-to-video generation.
\newblock \emph{CoRR}, abs/2305.14330, 2023{\natexlab{a}}.

\bibitem[Hong et~al.(2023{\natexlab{b}})Hong, Ding, Zheng, Liu, and Tang]{cogvideo}
Wenyi Hong, Ming Ding, Wendi Zheng, Xinghan Liu, and Jie Tang.
\newblock Cogvideo: Large-scale pretraining for text-to-video generation via transformers.
\newblock In \emph{The Eleventh International Conference on Learning Representations, {ICLR} 2023, Kigali, Rwanda, May 1-5, 2023}. OpenReview.net, 2023{\natexlab{b}}.

\bibitem[Huang et~al.(2023)Huang, Feng, Shi, Xu, Yu, and Yang]{free-bloom}
Hanzhuo Huang, Yufan Feng, Cheng Shi, Lan Xu, Jingyi Yu, and Sibei Yang.
\newblock Free-bloom: Zero-shot text-to-video generator with {LLM} director and {LDM} animator.
\newblock \emph{CoRR}, abs/2309.14494, 2023.

\bibitem[J{\"a}hne(2005)]{book1}
Bernd J{\"a}hne.
\newblock \emph{Digital image processing}.
\newblock Springer Science \& Business Media, 2005.

\bibitem[Khachatryan et~al.(2023)Khachatryan, Movsisyan, Tadevosyan, Henschel, Wang, Navasardyan, and Shi]{t2vz}
Levon Khachatryan, Andranik Movsisyan, Vahram Tadevosyan, Roberto Henschel, Zhangyang Wang, Shant Navasardyan, and Humphrey Shi.
\newblock Text2video-zero: Text-to-image diffusion models are zero-shot video generators.
\newblock \emph{CoRR}, abs/2303.13439, 2023.

\bibitem[Kim et~al.(2023)Kim, Baldi, and McAleer]{verify4}
Geunwoo Kim, Pierre Baldi, and Stephen McAleer.
\newblock Language models can solve computer tasks.
\newblock \emph{CoRR}, abs/2303.17491, 2023.

\bibitem[Li et~al.(2017)Li, Min, Shen, Carlson, and Carin]{early1}
Yitong Li, Martin~Renqiang Min, Dinghan Shen, David~E. Carlson, and Lawrence Carin.
\newblock Video generation from text.
\newblock \emph{CoRR}, abs/1710.00421, 2017.

\bibitem[Li et~al.(2019)Li, Gan, Shen, Liu, Cheng, Wu, Carin, Carlson, and Gao]{early2}
Yitong Li, Zhe Gan, Yelong Shen, Jingjing Liu, Yu Cheng, Yuexin Wu, Lawrence Carin, David~E. Carlson, and Jianfeng Gao.
\newblock Storygan: {A} sequential conditional {GAN} for story visualization.
\newblock In \emph{{IEEE} Conference on Computer Vision and Pattern Recognition, {CVPR} 2019, Long Beach, CA, USA, June 16-20, 2019}, pages 6329--6338. Computer Vision Foundation / {IEEE}, 2019.

\bibitem[Li et~al.(2023)Li, Liu, Wu, Mu, Yang, Gao, Li, and Lee]{gligen}
Yuheng Li, Haotian Liu, Qingyang Wu, Fangzhou Mu, Jianwei Yang, Jianfeng Gao, Chunyuan Li, and Yong~Jae Lee.
\newblock {GLIGEN:} open-set grounded text-to-image generation.
\newblock In \emph{{IEEE/CVF} Conference on Computer Vision and Pattern Recognition, {CVPR} 2023, Vancouver, BC, Canada, June 17-24, 2023}, pages 22511--22521. {IEEE}, 2023.

\bibitem[Lian et~al.(2023)Lian, Li, Yala, and Darrell]{lmd}
Long Lian, Boyi Li, Adam Yala, and Trevor Darrell.
\newblock Llm-grounded diffusion: Enhancing prompt understanding of text-to-image diffusion models with large language models.
\newblock \emph{CoRR}, abs/2305.13655, 2023.

\bibitem[Luo et~al.(2023)Luo, Chen, Zhang, Huang, Wang, Shen, Zhao, Zhou, and Tan]{videofusion}
Zhengxiong Luo, Dayou Chen, Yingya Zhang, Yan Huang, Liang Wang, Yujun Shen, Deli Zhao, Jingren Zhou, and Tieniu Tan.
\newblock Videofusion: Decomposed diffusion models for high-quality video generation.
\newblock In \emph{{IEEE/CVF} Conference on Computer Vision and Pattern Recognition, {CVPR} 2023, Vancouver, BC, Canada, June 17-24, 2023}, pages 10209--10218. {IEEE}, 2023.

\bibitem[OpenAI(2023)]{gpt4}
OpenAI.
\newblock {GPT-4} technical report.
\newblock \emph{CoRR}, abs/2303.08774, 2023.

\bibitem[Oppenheim et~al.(1979)Oppenheim, Lim, Kopec, and Pohlig]{phase2}
A Oppenheim, Jae Lim, Gary Kopec, and SC Pohlig.
\newblock Phase in speech and pictures.
\newblock In \emph{ICASSP'79. IEEE International Conference on Acoustics, Speech, and Signal Processing}, pages 632--637. IEEE, 1979.

\bibitem[Oppenheim and Lim(1981)]{phases3}
Alan~V Oppenheim and Jae~S Lim.
\newblock The importance of phase in signals.
\newblock \emph{Proceedings of the IEEE}, 69\penalty0 (5):\penalty0 529--541, 1981.

\bibitem[Piotrowski and Campbell(1982)]{phase4}
Leon~N Piotrowski and Fergus~W Campbell.
\newblock A demonstration of the visual importance and flexibility of spatial-frequency amplitude and phase.
\newblock \emph{Perception}, 11\penalty0 (3):\penalty0 337--346, 1982.

\bibitem[Radford et~al.(2021)Radford, Kim, Hallacy, Ramesh, Goh, Agarwal, Sastry, Askell, Mishkin, Clark, et~al.]{clip}
Alec Radford, Jong~Wook Kim, Chris Hallacy, Aditya Ramesh, Gabriel Goh, Sandhini Agarwal, Girish Sastry, Amanda Askell, Pamela Mishkin, Jack Clark, et~al.
\newblock Learning transferable visual models from natural language supervision.
\newblock In \emph{International conference on machine learning}, pages 8748--8763. PMLR, 2021.

\bibitem[Shinn et~al.(2023)Shinn, Cassano, Gopinath, Narasimhan, and Yao]{verify1}
Noah Shinn, Federico Cassano, Ashwin Gopinath, Karthik~R Narasimhan, and Shunyu Yao.
\newblock Reflexion: Language agents with verbal reinforcement learning.
\newblock In \emph{Thirty-seventh Conference on Neural Information Processing Systems}, 2023.

\bibitem[Singer et~al.(2023)Singer, Polyak, Hayes, Yin, An, Zhang, Hu, Yang, Ashual, Gafni, Parikh, Gupta, and Taigman]{makeavideo}
Uriel Singer, Adam Polyak, Thomas Hayes, Xi Yin, Jie An, Songyang Zhang, Qiyuan Hu, Harry Yang, Oron Ashual, Oran Gafni, Devi Parikh, Sonal Gupta, and Yaniv Taigman.
\newblock Make-a-video: Text-to-video generation without text-video data.
\newblock In \emph{The Eleventh International Conference on Learning Representations, {ICLR} 2023, Kigali, Rwanda, May 1-5, 2023}. OpenReview.net, 2023.

\bibitem[Weng et~al.(2023)Weng, Zhu, Xia, Li, He, Liu, and Zhao]{verify3}
Yixuan Weng, Minjun Zhu, Fei Xia, Bin Li, Shizhu He, Kang Liu, and Jun Zhao.
\newblock Large language models are better reasoners with self-verification, 2023.

\bibitem[Xie et~al.(2023)Xie, Ye, Li, Li, Lin, Zheng, Shen, and Shou]{visorgpt}
Jinheng Xie, Kai Ye, Yudong Li, Yuexiang Li, Kevin~Qinghong Lin, Yefeng Zheng, Linlin Shen, and Mike~Zheng Shou.
\newblock Visorgpt: Learning visual prior via generative pre-training.
\newblock \emph{CoRR}, abs/2305.13777, 2023.

\bibitem[Yilmaz et~al.(2023)Yilmaz, Lotman, Karjus, and Tikka]{camera1}
Mehmet~Burak Yilmaz, Elen Lotman, Andres Karjus, and Pia Tikka.
\newblock An embodiment of the cinematographer: emotional and perceptual responses to different camera movement techniques.
\newblock \emph{Frontiers in Neuroscience}, 17, 2023.

\bibitem[Zhang and Agrawala(2023)]{controlnet}
Lvmin Zhang and Maneesh Agrawala.
\newblock Adding conditional control to text-to-image diffusion models.
\newblock \emph{CoRR}, abs/2302.05543, 2023.

\bibitem[Zhang et~al.(2023)Zhang, Zhang, Vineet, Joshi, and Wang]{controlgpt}
Tianjun Zhang, Yi Zhang, Vibhav Vineet, Neel Joshi, and Xin Wang.
\newblock Controllable text-to-image generation with {GPT-4}.
\newblock \emph{CoRR}, abs/2305.18583, 2023.

\bibitem[Zhou et~al.(2022)Zhou, Wang, Yan, Lv, Zhu, and Feng]{MagicVideo}
Daquan Zhou, Weimin Wang, Hanshu Yan, Weiwei Lv, Yizhe Zhu, and Jiashi Feng.
\newblock Magicvideo: Efficient video generation with latent diffusion models.
\newblock \emph{CoRR}, abs/2211.11018, 2022.

\end{thebibliography}
}


\end{document}